\documentclass{article}
\usepackage{spconf,amsmath,graphicx,hyperref}
\usepackage{graphicx}
\usepackage{latexsym}
\usepackage{amssymb}
\usepackage{amsmath}
\usepackage{amsthm}
\usepackage{booktabs}
\usepackage{enumitem}
\usepackage{graphicx}
\usepackage{color}
\usepackage{amsfonts}   
\usepackage{booktabs}
\usepackage{algorithm}
\usepackage{algorithmic}
\usepackage{multirow}
\usepackage{multicol}
\usepackage{subcaption}
\usepackage{wasysym}
\usepackage{tabularx} 
\usepackage{booktabs} 
\usepackage[table]{xcolor}
\usepackage{url}

\usepackage{threeparttable}
\usepackage[toc,page]{appendix} 
\usepackage[switch]{lineno}

\title{AUTHOR GUIDELINES FOR ICASSP 2026 PROCEEDINGS MANUSCRIPTS}
\title{MCGA: Mixture of Codebooks Hyperspectral Reconstruction via Grayscale-Aware Attention}
\name{Zhanjiang Yang$^{1*}$, Lijun Sun$^{1*}$, Jiawei Dong$^1$, Xiaoxin An$^1$, Yang Liu$^2$, Meng Li$^{1\dagger
}$
\thanks{$^{*}$ Equal contribution.}
\thanks{$^{\dagger}$ Corresponding author.}
}
\address{$^1$Shenzhen Technology University, China\\
$^2$Swansea University, United Kingdom\\
2410263030@mails.szu.edu.cn, \{sunlijun, anxiaoxin, limeng2\}@sztu.edu.cn, yang.liu@swansea.ac.uk}
\begin{document}
%
\maketitle
\begin{abstract}
Reconstructing hyperspectral images (HSIs) from RGB inputs provides a cost-effective alternative to hyperspectral cameras, but reconstructing high-dimensional spectra from three channels is inherently ill-posed. Existing methods typically directly regress RGB-to-HSI mappings using large attention networks, which are computationally expensive and handle ill-posedness only implicitly. We propose MCGA, a Mixture-of-Codebooks with Grayscale-aware Attention framework that explicitly addresses these challenges using spectral priors and photometric consistency. MCGA first learns transferable spectral priors via a mixture-of-codebooks (MoC) from heterogeneous HSI datasets, then aligns RGB features with these priors through grayscale-aware photometric attention (GANet). Efficiency and robustness are further improved via top-$K$ attention design and test-time adaptation (TTA). Experiments on multiple real-world benchmarks demonstrate the state-of-the-art accuracy, strong cross-dataset generalization, and 4–5$\times$ faster inference. 
Codes will be available once acceptance at \url{https://github.com/Fibonaccirabbit/MCGA}.
\end{abstract}
\section{Introduction}
\label{sec:intro}
Hyperspectral images (HSIs) capture dozens to hundreds of contiguous spectral bands with sub-10 nm resolution~\cite{bian2024broadband}, providing richer material and structural information than multispectral or RGB images. This enables applications in medicine~\cite{zhang2024medical}, agriculture~\cite{ahmed2025comprehensive}, land cover classification~\cite{lou2025land}, and target detection~\cite{lei2023a}. However, hyperspectral cameras are expensive and slow, scanning one spectral band or spatial line at a time, which limits real-time deployment.

Learning-based RGB-to-HSI reconstruction offers a promising solution~\cite{zhang2022survey,chen2023review}. Attention-based methods (MST++~\cite{mst++}, HRNet~\cite{hrnet}, GMSR~\cite{gmsr}, R3ST~\cite{r3st}) achieve high accuracy but are computationally heavy, while residual/dense networks (HSCNN+~\cite{HSCNN}, AGDNet~\cite{AGDNet}) generalize poorly under variations in illumination, sensor response, or noise. As a result, existing approaches struggle with both efficiency and robustness, limiting their practicality for real-world applications.


RGB-to-HSI reconstruction is inherently \textbf{ill-posed} due to the extreme dimensionality gap. It recovers narrow-band spectra from broad-band RGB, posing a spectral augmentation problem with strict photometric requirements, where even small pixel errors can affect physical fidelity. This is different from RGB super-resolution, which maps RGB to RGB and emphasizes visual consistency. 
Existing methods often \textbf{directly} learn the RGB-to-HSI mapping with over-parameterized models; however, although RGB and HSI share semantic structure, they differ primarily in grayscale intensity across bands, motivating grayscale-aware modeling.

\begin{figure*}[t]
  \centering
   \includegraphics[width=0.9\linewidth]{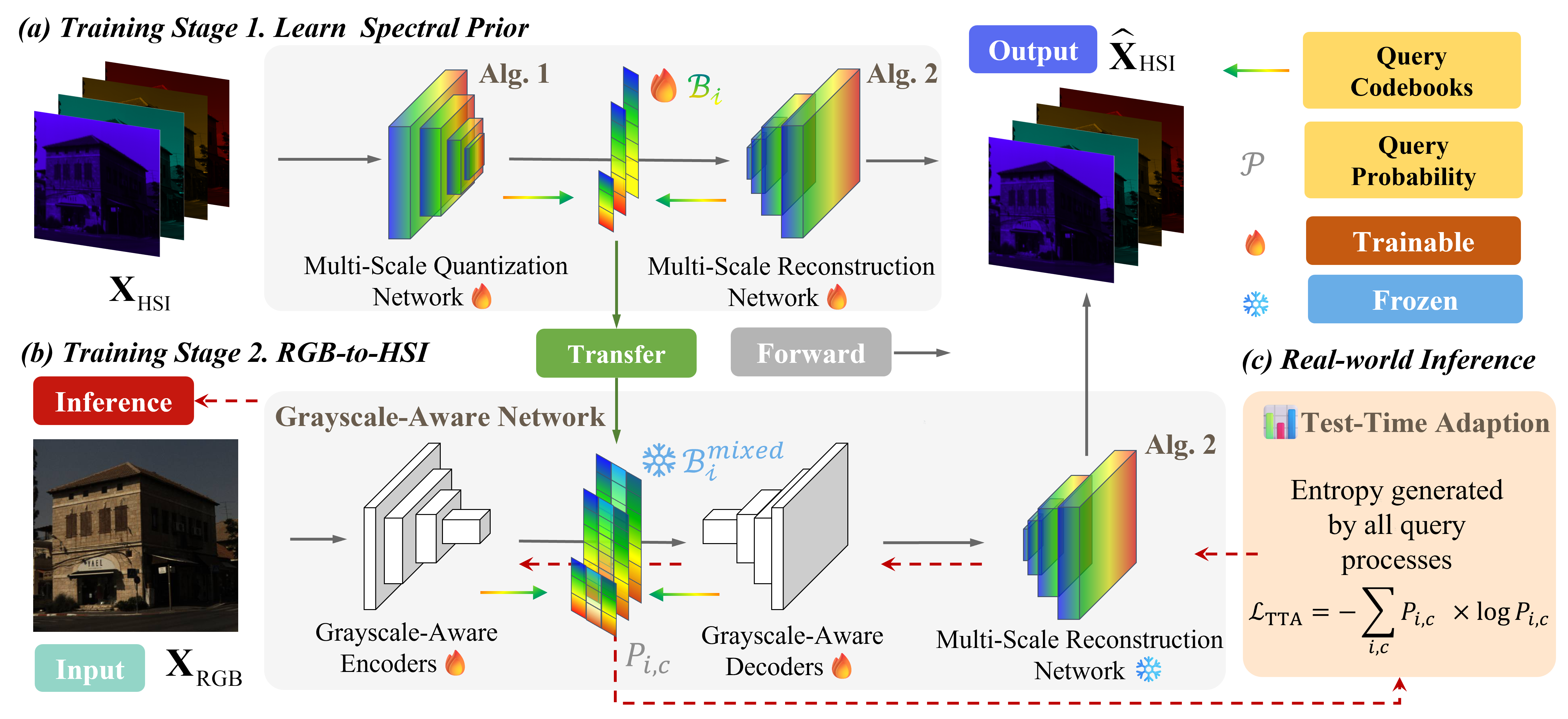}
   \caption{The proposed MCGA is a Mixture-of-Codebooks framework with Grayscale-aware Attention, leveraging spectral priors and grayscale photometric consistency.}
   \label{fig:network}
\end{figure*}

We propose \textbf{MCGA}, a \textbf{M}ixture-of-\textbf{C}odebooks with \textbf{G}rayscale-aware \textbf{A}ttention framework that addresses the ill-posed RGB-to-HSI problem via spectral priors and photometric consistency (Fig.~\ref{fig:network}). Rather than directly learning the RGB-to-HSI mapping, MCGA adopts a two-stage paradigm.  \textbf{Stage~1}:, transferable spectral priors are learned as a \textbf{mixture of codebooks} from heterogeneous HSI datasets by a multi-scale vector-quantized variational autoencoder (VQ-VAE). \textbf{Stage~2}: a grayscale-aware attention network (GANet) aligns RGB features to these priors and captures spectral intensity variations efficiently.
A \textbf{top $K$ attention mechanism} reduces complexity from $\mathcal{O}(C^2HW)$ to $\mathcal{O}(C^2K)$ with minimal accuracy loss, enabling $4$–$5\times$ faster inference. Finally, a \textbf{test-time adaptation} strategy further improves robustness under varying illumination and sensor responses.
\textbf{Contributions:}
\begin{itemize}
\item \textbf{MCGA}: A two-stage RGB-to-HSI framework that embeds physical signal constraints via mixture-of-codebooks priors and RGB-aligned modeling.
\item \textbf{GANet}: A grayscale-aware attention network with top-$K$ attention and test-time adaptation for efficient and robust spectral modeling.
\item \textbf{Evaluation}: Experiments demonstrate state-of-the-art accuracy, generalization, and real-time performance.
\end{itemize}

\section{Proposed Method}

\subsection{Problem Statement}
HSI reconstruction aims to recover a spectral cube $\mathbf{X}_\textrm{HSI}\in\mathbb{R}^{C\times H\times W}$ from an RGB image $\mathbf{X}_\textrm{RGB}\in\mathbb{R}^{3\times H\times W}$, where $C$, $H$, $W$ denote spectral channels, height, and width, respectively. Formally, the objective is to learn: $\mathbf{X}_\textrm{RGB}\rightarrow \mathbf{X}_\textrm{HSI}$.

\subsection{Stage~1: Spectral Prior via Multi-Scale VQ-VAE}

RGB cues are insufficient for RGB-to-HSI reconstruction, motivating a  spectral prior. We derive this prior from a self-supervised VQ-VAE trained on HSIs, where dataset-specific codebooks are concatenated into a transferable Mixture of Codebooks (MoC). 
The MoC captures dataset diversity and cross-dataset spectral variability, yielding a universal prior.

\begin{algorithm}[htbp]
   \caption{Multi-scale Quantization (in Stage 1)}
   \label{alg:msq}
    \begin{algorithmic}[1]
       \STATE {\bf In:} $\mathbf{X}_\textrm{HSI}$, scales $S$, $\beta$; {\bf Out:} $\mathcal{B}, \mathcal{H}^d=\{\mathbf{H}_i^d\}_{i=1}^S, \mathcal{L}_1$
       \STATE $\mathcal{L}_1=0$, $C_m=2^{\lfloor \log_2^{\frac{C}{2}}\rfloor}$
              {\scriptsize \hfill {\color{blue!60}{$\triangleright$ Approximately half spectral channels}}}
       \STATE $f=\text{SpectralMask}(\mathbf{X}_\textrm{HSI})$ 
              {\scriptsize \hfill {\color{blue!60}{$\triangleright$ Random mask $C_m$ spectral channels}}}
       \STATE $\mathcal{B}=\{\mathcal{B}_i\sim\mathcal{N}(0,I)\in\mathbb{R}^{512\times \frac{C_m}{2^i}}\}_{i=1}^S$
              {\scriptsize \hfill {\color{blue!60}{$\triangleright$ Spectral codebooks}}}
       \FOR{$i=1$ {\bf to} $S$}
          \STATE $\mathbf{H}_i^d=\text{Downsample}(f)$ 
                 {\scriptsize \hfill {\color{blue!60}{$\triangleright$ Downsample spatially}}}
          \STATE $\mathbf{H}_i^q=\text{Quantize}(\mathcal{B}_i,\mathbf{H}_i^d)$ 
                 {\scriptsize \hfill {\color{blue!60}{$\triangleright$ Quantize to  nearest codeword in $\mathcal{B}_i$}}}
          \STATE $\mathcal{L}_1 {+}{=} \mathcal{L}_{\text{embed}}(\mathbf{H}_i^d,\mathbf{H}_i^q)+\beta\mathcal{L}_{\text{commit}}(\mathbf{H}_i^d,\mathbf{H}_i^q)$
                 {\scriptsize \hfill {\color{blue!60}{$\triangleright$ Sec. 2.5}}}
          \STATE $f=\phi(\text{Upsample}(\mathbf{H}_i^q))$ 
                 {\scriptsize \hfill {\color{blue!60}{$\triangleright$ Upsample and convolutionally fuse}}}
       \ENDFOR
    \end{algorithmic}
\end{algorithm}

\begin{algorithm}[htbp]
   \caption{Multi-scale Reconstruction (in Stage 1)}
   \label{alg:msr}
    \begin{algorithmic}[1]
       \STATE {\bf In:} $\mathcal{B}=\{\mathcal{B}_i\}_{i=1}^S$, $\mathcal{H}^d=\{\mathbf{H}_i^d\}_{i=1}^S$, $\beta$; \ {\bf Out:} $\hat{\mathbf{X}}_\textrm{HSI}, \mathcal{L}_2$
       \STATE $f=0$, $\mathcal{L}_2=0$, $S=|\mathcal{\mathcal{B}}|$
              {\scriptsize \hfill {\color{blue!60}{$\triangleright$ Initialize}}}
       \FOR{$i=S$ {\bf down to} $1$}
          \STATE $\mathbf{R}_i^q=\text{Quantize}(\mathcal{B}_i,\mathbf{H}_i^d)$
                 {\scriptsize \hfill {\color{blue!60}{$\triangleright$ Quantize to  nearest codeword in $\mathcal{B}_i$}}}
          \STATE $\mathcal{L}_2 {+}{=} \mathcal{L}_{\text{embed}}(\mathbf{R}_i^q,\mathbf{H}_i^d)+\beta\mathcal{L}_{\text{commit}}(\mathbf{R}_i^q,\mathbf{H}_i^d)$
                 {\scriptsize \hfill {\color{blue!60}{$\triangleright$ Sec. 2.5}}}
          \STATE $f=\text{Concat}(f,\phi(\text{Upsample}(\text{Concat}(\mathbf{H}_i^d,\mathbf{R}_i^q))))$ \\
                 {\scriptsize \hfill {\color{blue!60}{$\triangleright$ $\text{Concat}(\cdot)$ means concatenation}}}
       \ENDFOR
       \STATE $\hat{\mathbf{X}}_\textrm{HSI}=\phi(f)$ 
              {\scriptsize \hfill {\color{blue!60}{$\triangleright$ Convolutionally fuse}}}
    \end{algorithmic}
\end{algorithm}

As shown in Fig.~\ref{fig:network}(a), the quantization network (Alg.~\ref{alg:msq}) downsamples $\mathbf{X}_\textrm{HSI}$ features, discretizes them into scale-specific codebooks $\mathcal{B}=\{\mathcal{B}_i\}_{i=1}^S$, and reinjects quantized representations $\mathcal{H}^d=\{\mathbf{H}_i^d\}_{i=1}^S$. The reconstruction network (Alg.~\ref{alg:msr}) then upsamples and fuses them to recover $\hat{\mathbf{X}}_\textrm{HSI}$. 
These two networks are trained using the loss functions defined in Sec.~\ref{sec:loss}.
By independently training on $N$ HSI datasets $\{\mathbf{X}_\textrm{HSI}^{(n)}\}_{n=1}^N$ and deriving the corresponding codebooks $\mathcal{B}_i^{(n)}$, their concatenation yields the mixture of codebooks $\mathcal{B}^{\text{mixed}} = \{\mathcal{B}_i^{\text{mixed}} = \big[\, \mathcal{B}_i^{(n)} \,\big]_{n=1}^{N}\}_{i=1}^S$.

\subsection{Stage 2: RGB-to-HSI via Grayscale-Aware Network}
As shown in Fig.~\ref{fig:network}(b), in Stage~2, we introduce a grayscale-aware network (GANet) for RGB-to-HSI reconstruction. GANet first aligns RGB features from the input $\mathbf{X}_\textrm{RGB}$ with the MoC priors $\mathcal{B}^{\textrm{mixed}}$ learned in Stage~1, and then transforms the aligned features into HSI representations $\hat{\mathbf{X}}_\textrm{HSI}$ via Alg.~\ref{alg:msr}.


GANet adopts a Transformer-based encoder–decoder architecture with grayscale-aware (GA) operations embedded in the self-attention and feedforward modules, forming grayscale-aware Transformer blocks (GABs) to model intensity variations for photometric consistency.

\textbf{Grayscale-Aware Operations.}
We design a learnable gamma correction  $\textrm{GA}_\gamma$ for  brightness control and a learnable logarithmic transformation $\textrm{GA}_l$ for grayscale adjustment. 
\begin{equation}
   \textrm{GA}_\gamma(\mathbf{X})=\mathbf{X}^{\mathbf{a}}, \quad 
   \textrm{GA}_l(\mathbf{X})=(1+4\mathbf{a})\log(1+\mathbf{X})
\end{equation}
where awareness vector $\mathbf{a}$ is derived by a softmax-activated MLP on the spatial global average of normalized $X$.

\textbf{Grayscale-Aware Self-Attention.}
Conventional self-attention mechanisms compute the attention across all spectral channels, incurring $\mathcal{O}(C^2HW)$ complexity.
In contrast, we compute the query and key using the top $K$ representative quantized vectors selected based on codebook hit rates, reducing the complexity to $\mathcal{O}(C^2K)$ ($K\ll HW$).
Thus, $Q_i=\mathrm{GA}_l(\mathrm{Quantize}_K(\mathcal{B}_i^{\text{mixed}},\, \mathbf{X}))$, $K_i=\mathrm{Quantize}_K(\mathcal{B}_i^{\text{mixed}},\, \mathbf{X})$, and $V_i=\mathbf{X}$, where $\mathrm{Quantize}_K(\cdot)$ denotes the top $K$ outputs of  $\mathrm{Quantize}(\cdot)$, and $\mathrm{GA}_l$ is applied to enhance grayscale awareness while ensuring stable training.

\textbf{Grayscale-Aware Feedforward.}
The module input $\mathbf{X}$ is processed by a series of convolutional operations to produce $\mathbf{X}_1$ with doubled spectral channels.
$\mathbf{X}_1$ is then evenly split along the spectral dimension into $\mathbf{X}_2$ and $\mathbf{X}_3$.
The final module output is obtained by applying convolutional operations to the concatenation $\mathrm{Concat}(\mathbf{X}_1, \mathrm{GA}_{\gamma}(\mathbf{X}_2), \mathrm{GA}_l(\mathbf{X}_3))$, enabling the model to adaptively control the contribution of GA-processed features.

\subsection{Real-World Inference}
In real-world deployment, RGB inputs often deviate from training distributions due to illumination, sensor, or scene shifts, degrading reconstruction performance. Under fixed spectral priors, GANet may misalign RGB features with the MoC. To address this without labels, we employ lightweight test-time adaptation (TTA) by minimizing the entropy of MoC assignments, \(\mathcal{L}_{\text{TTA}}=-\sum_{i,c} P_{ic}\log P_{ic}\), where \(P_{ic}\in\mathbb{R}^{HW\times512}\) denotes the probabilistic assignment matrix of codebooks. Only the affine parameters of GA attention are updated, while the encoder, MoC, and decoder are frozen, enabling efficient feature realignment and robust hyperspectral reconstruction under distribution shifts.

\subsection{Loss Function}\label{sec:loss}
Stage~1 and Stage~2 optimize $\mathcal{L}_{S1}=\mathcal{L}_\textrm{rec}+\beta(\mathcal{L}_1+\mathcal{L}_2)$ and $\mathcal{L}_{S2}=\mathcal{L}_\textrm{rec}$, respectively,  
where $\mathcal{L}_\textrm{rec}=\sqrt{(\mathbf{X}_\textrm{HSI}-\hat{\mathbf{X}}_\textrm{HSI})^2+\epsilon}$ with $\epsilon=10^{-6}$~\cite{lai2018fast} is the reconstruction loss,
and $\mathcal{L}_1$ and $\mathcal{L}_2$ (Alg.~1-2) aggregate the embedding and commitment losses $\mathcal{L}_\textrm{embed}=\|sg[\mathbf{X}_q]-\mathbf{X}\|_2^2$  and $\mathcal{L}_\textrm{commit}=\|\mathbf{X}_q-sg[\mathbf{X}]\|_2^2$ with $sg(\cdot)$ denoting the stop gradient operator.

\section{Experiments}
\subsection{Datasets}
\label{subsec:datasets} 
We evaluate on two large-scale real-world RGB–HSI benchmarks. HySpecNet-11k~\cite{fuchs2023hyspecnet} contains 11k $128\times128$ RGB–HSI patches with 224 bands (420–2450 nm) and provides \emph{easy} and \emph{hard} splits; to avoid leakage, we use only the hard split (8k/2k/1k for train/val/test). ARAD-1k~\cite{arad2022ntire} consists of 1k $482\times512$ images with 31 bands (400–700 nm, 10 nm interval); following the official protocol, images are cropped to $128\times128$ patches with 900 training and 50 validation samples, while the remaining 50 test images are kept private.

To construct the MoC, we additionally use HyperGlobal-450K~\cite{hyperglobal} with 1,701 $64\times64$ HSI samples of 191 bands.

\subsection{Baselines}
To evaluate the proposed MCGA, we compare it with two categories of state-of-the-art methods: 
(i) attention-based approaches, including MST++~\cite{mst++}, HRNet~\cite{hrnet}, GMSR~\cite{gmsr}, and R3ST~\cite{r3st}; and 
(ii) residual/dense network–based methods, including HSCNN+~\cite{HSCNN} and AGDNet~\cite{AGDNet}.
MCGA-S2 denotes the variant employing two-scale feature extraction.

\subsection{Implementation}
All experiments use NVIDIA V100 GPU with 32GB memory. 

\textbf{Hyperparameters.} AdamW is used with a learning rate of $4\times10^{-4}$ and CycleScheduler~\cite{smith2017cyclical}; \(\beta=0.25\)~\cite{razavi2019generating}.

\textbf{Metrics.} Following the NTIRE2022 Spectral Reconstruction Challenge, we report the root mean square error (RMSE) and mean relative absolute error (MRAE) as primary metrics, where $\textrm{MRAE}(Y,\hat{Y})=\frac{1}{N}\sum_{i=1}^{N}\frac{|Y_i-\hat{Y}_i|}{Y_i}$ represents the mean pixel-wise percentage error. Peak signal-to-noise ratio (PSNR) is included as a supplementary metric.

\begin{table*}[t]
    \centering
    \caption{
        Accuracy--efficiency trade-off on ARAD-1k and HySpecNet-11k. 
        ``Params'' = model size (M); ``Time'' = inference time per image (ms). 
        MCGA-S2 achieves state-of-the-art accuracy and 4--5$\times$ speedup over the second best R3ST.
    }
    \label{tab:arad_results}
    \renewcommand{\arraystretch}{1.15}
    \setlength{\tabcolsep}{4pt}
    \resizebox{\linewidth}{!}{
    \begin{tabular}{lcc|ccc|ccc|ccc|ccc}
        \toprule
        \textbf{Method} & \textbf{Params} & \textbf{Time} 
        & \multicolumn{3}{c|}{\textbf{ARAD-1k (val)}} 
        & \multicolumn{3}{c|}{\textbf{ARAD-1k (val, mixed)}} 
        & \multicolumn{3}{c|}{\textbf{HySpecNet-11k (test)}} 
        & \multicolumn{3}{c}{\textbf{HySpecNet-11k (test, mixed)}} \\
        \cmidrule{4-15}
        & (M) & (ms) & RMSE↓ & MRAE↓ & PSNR↑ & RMSE↓ & MRAE↓ & PSNR↑ & RMSE↓ & MRAE↓ & PSNR↑ & RMSE↓ & MRAE↓ & PSNR↑ \\
        \midrule
        HSCNN+~\cite{HSCNN}      & 4.65  & 246.03 & 0.0588 & 38.1 & 26.39 & 0.0939 & 48.3 & 22.37 & 0.0279 & 19.6 & 33.36  & 0.0336 & 21.9 & 31.51 \\
        HRNet~\cite{hrnet}       & 31.70 & 381.19 & 0.0550 & 34.8 & 26.89 & 0.0745 & 41.0 & 24.23 & 0.0330 & 22.3 & 31.43  & 0.0343 & 23.0 & 31.14 \\
        AGDNet~\cite{AGDNet}      & \textbf{0.17}  & \underline{112.12} & 0.0473 & 42.4 & 27.42 & 0.0635 & 59.9 & 24.36 & 0.0248 & 17.2 & 33.95 & 0.0266 & 18.2 & 33.20 \\
        GMSR\cite{gmsr}        & \underline{0.20}  & 460.38 & 0.0495 & 33.9 & 28.18 & 0.0858 & 43.9 & 23.83 & 0.0283 & 19.7 & 32.83 & 0.0301 & 20.5 & 32.35 \\
        R3ST~\cite{r3st}        & 1.64  & 441.22 & 0.0266 & 19.6 & 33.48 & \underline{0.0414} & \underline{27.2} & \underline{29.21} & \underline{0.0210} & \underline{15.4} & \underline{35.83} & \underline{0.0252} & \underline{17.5} & \underline{34.19} \\
        MST++~\cite{mst++}       & 1.62 & 435.76 & \underline{0.0248} & \underline{16.5} & \underline{34.32} & 0.0671 & 47.9 & 25.00 & 0.0222 & 15.7 & 35.31  & 0.0253 & 17.6 & 34.10 \\
        \rowcolor{gray!8}
        \textbf{MCGA-S2 (Ours)} & 0.76 & \textbf{93.60} & \textbf{0.0182} & \textbf{13.1} & \textbf{36.18} & \textbf{0.0319} & \textbf{20.9} & \textbf{31.26} & \textbf{0.0183} & \textbf{13.8} & \textbf{36.56} & \textbf{0.0208} & \textbf{15.1} & \textbf{35.63} \\
        \bottomrule
    \end{tabular}
    }
\end{table*}

\begin{table*}[t]
    \centering
    \footnotesize
    \begin{minipage}{0.48\linewidth}
        \centering
        \caption{Performance under $\pm10\%$ illumination perturbations on ARAD-1k.}
        \label{tab:tta_brightness}
        \setlength{\tabcolsep}{3pt}
        \renewcommand{\arraystretch}{1.1}
        \begin{tabular}{l|ccc|ccc}
            \toprule
            \multirow{2}{*}{\textbf{Method}} 
            & \multicolumn{3}{c|}{\textbf{+10\%}} 
            & \multicolumn{3}{c}{\textbf{-10\%}} \\
            & RMSE↓ & MRAE↓ & PSNR↑ & RMSE↓ & MRAE↓ & PSNR↑ \\
            \midrule
            MST++ & 0.0575 & 41.6 & 27.0 & 0.0763 & 76.4 & 24.5 \\
            MCGA-S2 & 0.0240 & 22.2 & 33.2 & 0.0327 & 38.3 & 30.7 \\
            \textbf{MCGA-S2+TTA} & \textbf{0.0227} & \textbf{17.7} & \textbf{34.0} & \textbf{0.0273} & \textbf{28.8} & \textbf{31.2} \\
            \bottomrule
        \end{tabular}
    \end{minipage}
    \hfill
    \begin{minipage}{0.48\linewidth}
        \centering
        \caption{Component-wise ablation results using MRAE$\downarrow$.}
        \label{tab:ablation}
        \setlength{\tabcolsep}{8pt}
        \renewcommand{\arraystretch}{1.2}
        \begin{tabular}{lcc}
            \toprule
            \textbf{Component} & \textbf{HySpecNet-11k (val)} & \textbf{ARAD-1k (val)} \\
            \midrule
            Plain GANet & 50.1 & 46.3 \\
            + Mixture of Codebooks & 34.0 \scriptsize{(-16.1\%)} & 29.8 \scriptsize{(-16.5\%)} \\
            + GA & 23.5 \scriptsize{(-10.5\%)} & 18.4 \scriptsize{(-11.4\%)} \\
            + Quantized Attn & 20.8 \scriptsize{(-2.7\%)} & 13.1 \scriptsize{(-5.3\%)} \\
            \midrule
            \quad$\circlearrowleft$ Single Codebook & 23.6 \scriptsize{(+3.8\%)} & 15.2 \scriptsize{(+2.1\%)} \\
            \quad$\circlearrowleft$ Full Attn & 19.6 \scriptsize{(-1.2\%)} & 12.4 \scriptsize{(-0.7\%)} \\
            \bottomrule
        \end{tabular}
    \end{minipage}
\end{table*}


\subsection{State-of-the-Art Spectral Reconstruction}
\begin{figure}[ht]
  \centering
   \includegraphics[width=0.48\textwidth]{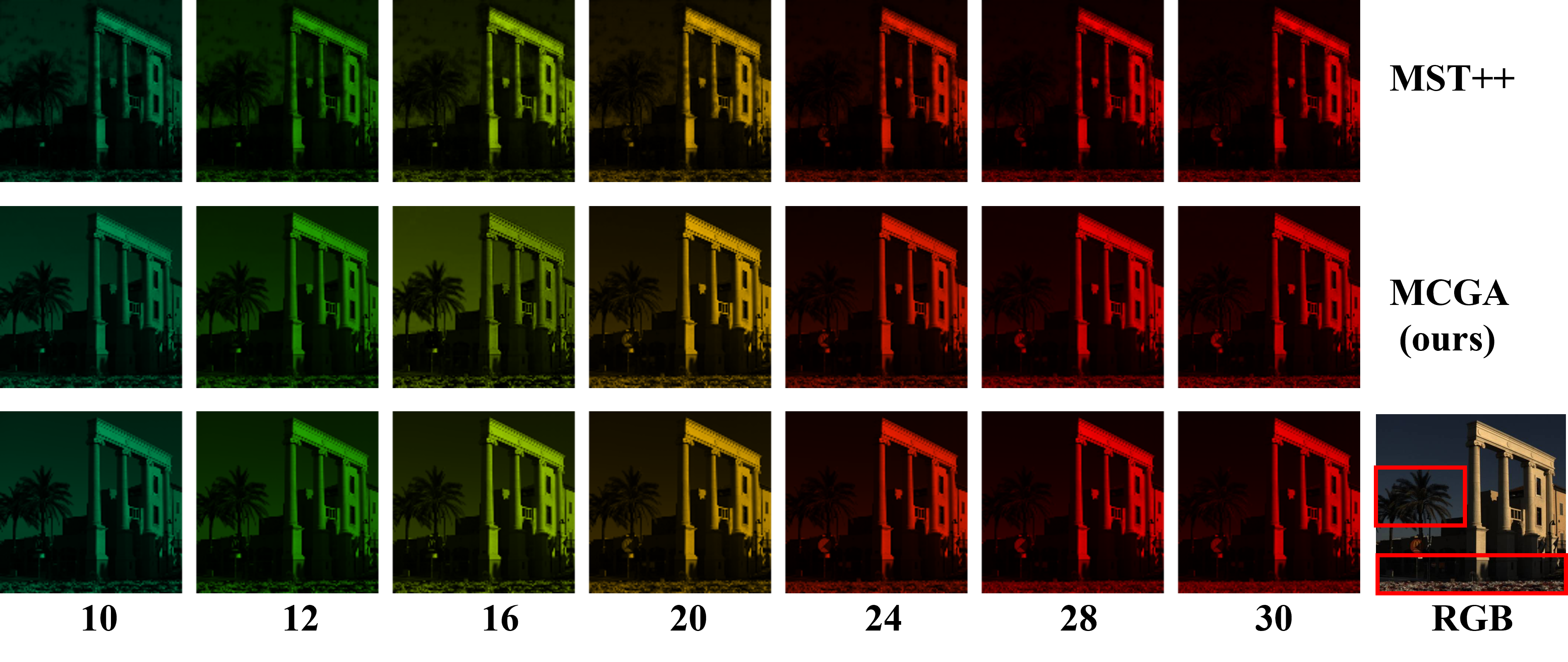}
   \caption{A case study on ARAD-1k: the bottom row shows ground truths for each channel, indicated by the numbers.}
   \label{fig:visualize}
\end{figure}

Table~\ref{tab:arad_results} reports results on ARAD-1k and HySpecNet-11k. 
With \(S=2\) and top \(K=16^2\), MCGA-S2 achieves state-of-the-art accuracy with substantially higher efficiency. 
On ARAD-1k, it reduces RMSE/MRAE by 27\%/3\% over MST++, improves PSNR by 5\%, and achieves a \(4.6\times\) speedup. 
On HySpecNet-11k, MCGA outperforms R3ST with 13\% lower RMSE, 1.6\% lower MRAE, and a \(5\times\) faster runtime. 
Fig.~\ref{fig:visualize} shows qualitative comparisons on ARAD-1k.

\subsection{Robustness to Spatial OOD}


In the \textbf{mixed setting} (labeled as ``mixed'' in Table~\ref{tab:arad_results}), RGB patches are randomly shuffled, preserving spectral cues but destroying spatial layouts, i.e., out-of-distribution (OOD) spatial structures. On HySpecNet-11k, all methods remain stable, while on ARAD-1k convolution-based methods collapse (e.g., MST++ RMSE: 0.0248→0.0671). MCGA-S2 achieves the best accuracy and degrades only mildly, showing that \textbf{pixel-level MoC encoding} and \textbf{grayscale-aware attention} enable robust generalization beyond spatial correlations.

\subsection{Robustness to Illumination Perturbations}
We simulate illumination (distribution) shifts on ARAD-1k using $\gamma$-correction ($\gamma\!=\!0.9,1.1$), as reported in Table~\ref{tab:tta_brightness}. All methods degrade notably under reduced illumination ($\gamma\!=\!0.9$), where spectral cues are weakened. With test-time adaptation (TTA), MCGA substantially improves robustness, achieving $\sim$10\% lower MRAE compared to its non-TTA variant. These results highlight the effectiveness of entropy-minimization TTA in realigning RGB features to the MoC manifold under distribution shifts.


\section{Ablation Study}

\textbf{Scaling.} Performance peaks at multi-scale $S=2$; larger values overfit and degrade upsampling. For quantized attention, top $K=16^2$ gives the best accuracy–efficiency trade-off.

\textbf{Component-wise analysis.} 
We perform component-level ablation studies under the condition \(S=2\), as summarized in Table~\ref{tab:ablation}. ``Plain GANet" denotes the baseline architecture that retains only the convolutional operators and activation functions of GANet, with an unmodified computational flow; core components are then incrementally incorporated. In the full GANet configuration, we additionally evaluate substituting the Mixture of Codebooks (MoC) with homogeneous codebooks and replacing quantized self-attention with standard full self-attention.  


\section{Conclusion}
We presented MCGA, a two-stage framework for RGB-to-HSI reconstruction with strong robustness to real-world distribution shifts. Stage~1 learns transferable spectral priors via a multi-scale VQ-VAE, yielding a Mixture of Codebooks (MoC) that captures cross-dataset diversity. Stage~2 employs a lightweight Grayscale-Aware Network (GANet) to align RGB features with MoC, while top-$K$ attention significantly reduces complexity and test-time adaptation (TTA)  further improves resilience under photometric perturbations. Beyond spectral reconstruction, MoC offers potential for synthetic HSI generation, and the grayscale-aware attention can be extended to broader low-quality image recovery tasks.

\vfill\pagebreak

\bibliographystyle{IEEEbib}
\bibliography{strings,refs}

\section{Additional Results}

\begin{table}[!h]
\centering
\captionsetup{width=\linewidth}
  \caption{Performance comparison on the HySpecNet-11k mixed  dataset.}
    \renewcommand{\arraystretch}{1.2}
     \resizebox{\textwidth}{!}{
    \begin{tabular}{ccccccccccccccc}
    \toprule 
     \multirow{2}{*}{Models}& \multirow{2}{*}{Params}& \multirow{2}{*}{Times}&\multicolumn{3}{c}{HySpec11k Valid $32\times 32$} & \multicolumn{3}{c}{HySpec11k Valid $8\times 8$}& \multicolumn{3}{c}{HySpec11k Test $32\times 32$} & \multicolumn{3}{c}{HySpec11k Test $8\times 8$} \\ \cline{4-15}
    &&&RMSE & MRAE &PSNR &RMSE &MRAE &PSNR &RMSE &MRAE &PSNR &RMSE &MRAE &PSNR    \\
    \midrule 
    \rowcolor{gray!10}HSCNN+\cite{HSCNN}&7.65M&300.49ms& 0.0351&26.3\%&31.35&0.0428&29.9\%&29.73&0.0310&20.6\%&32.15&0.0378&24.6\%&30.56\\

    HRNet\cite{hrnet}&31.8M&469.17ms&0.0381&30.4\%&30.00&0.0405&31.9\%&29.50&0.0336&22.6\%&31.29&0.0356&23.8\%&30.84\\

    \rowcolor{gray!10}AGDNet\cite{AGDNet}&6.8M&981.24 ms&0.0290&25.5\%&32.80&0.0323&26.9\%&31.82&0.0256&17.6\%&33.57&\underline{0.0295}&\underline{19.9\%}&\underline{32.76}\\

    GMSR\cite{gmsr}&0.30M&544.42 ms&0.0316&28.0\%&31.96&0.0324&28.2\%&31.71&0.0292&20.0\%&32.57&0.0315&21.1\%&32.07\\

    \rowcolor{gray!10}MST++\cite{mst++}&1.67M&536.76 ms&0.0251&22.7\%&34.39&0.0313&25.5\%&32.48&0.0230&16.1\%&34.96&0.0310&20.1\%&32.53\\

    R3ST\cite{r3st}&\underline{1.79M}&\underline{592.61 ms}&\underline{0.0249}&\underline{22.5\%}&\underline{34.49}&\underline{0.0308}&\underline{25.2\%}&\underline{32.60}&\underline{0.0221}&\underline{15.9\%}&\underline{35.28}&0.0314&20.9\%&32.48\\

    \rowcolor{gray!10}MCGA-S2&0.80M&\textbf{117.41 ms}&\textbf{0.0207}&\textbf{19.2\%}&\textbf{35.44}&\textbf{0.0262}&\textbf{23.9\%}&\textbf{33.69}&\textbf{0.0193}&\textbf{14.2\%}&\textbf{36.20}&\textbf{0.0264}&\textbf{18.5\%}&\textbf{34.03}\\
    
    Improvement&\textcolor[HTML]{0071BB}{-\%}&\textcolor[HTML]{0071BB}{-\%}&\textcolor[HTML]{0071BB}{-17\%}&\textcolor[HTML]{0071BB}{-1.3\%}&\textcolor[HTML]{0071BB}{+3\%}&\textcolor[HTML]{0071BB}{-15\%}&\textcolor[HTML]{0071BB}{-1.3\%}&\textcolor[HTML]{0071BB}{+3\%}&
    \textcolor[HTML]{0071BB}{-13\%}
    &\textcolor[HTML]{0071BB}{-1.7\%}&\textcolor[HTML]{0071BB}{+3\%}&\textcolor[HTML]{0071BB}{-11\%}&\textcolor[HTML]{0071BB}{-1.4\%}&\textcolor[HTML]{0071BB}{+4\%}\\
    \bottomrule 
    \label{cmp:hyspec} 
    \end{tabular}}
\end{table}
\begin{table}[!h]
\centering
 \caption{Performance comparison on the ARAD-1k mixed dataset.}
    \renewcommand{\arraystretch}{1.2}
     \resizebox{0.8\textwidth}{!}{
    \begin{tabular}{ccccccccc}
    \toprule 
     \multirow{2}{*}{Model}& \multirow{2}{*}{Param.}& \multirow{2}{*}{Time} &\multicolumn{3}{c}{ARAD-1K Valid $32\times 32$ mixed} & \multicolumn{3}{c}{ARAD-1K Valid $8\times 8$ Mixed}\\ \cline{4-9}
    &&& RMSE & MRAE & PSNR &RMSE & MRAE & PSNR    \\
    \midrule 
    \rowcolor{gray!10}HSCNN+\cite{HSCNN}&4.65M&246.03 ms&
    0.0824&41.5\%&23.12&0.1157&56.1\%&20.80\\

    HRNet\cite{hrnet}&31.70M&381.19 ms&
    0.0750&37.3\%&24.29&0.1084&54.5\%&21.14\\

    \rowcolor{gray!10}AGDNet\cite{AGDNet}&0.17M&112.12 ms&0.0616&50.8\%&24.73&0.0905&92.2\%&21.59\\
    
    GMSR\cite{gmsr}&0.20M&460.38 ms&0.0804&39.0\%&24.41&0.1045&51.2\%&22.08\\
    
    \rowcolor{gray!10}MST++\cite{mst++}&1.67M&536.76 ms&0.0648&39.3\%&	25.80&0.0846&65.1\%&22.93\\
    
    R3ST\cite{r3st}&1.64M&441.22 ms&\underline{0.0364}&\underline{22.4\%}&\underline{30.45}&\underline{0.0638}&\underline{38.5\%}&\underline{25.82}\\
    
    \rowcolor{gray!10}
    MCGA-S2&0.76M&\textbf{93.60 ms}&\textbf{0.0305}&\textbf{17.6\%}&\textbf{31.88}&\textbf{0.0538}&\textbf{35.3\%}&\textbf{27.00}\\

    Improvement&\textcolor[HTML]{0071BB}{-\%}&\textcolor[HTML]{0071BB}{-\%}&\textcolor[HTML]{0071BB}{-16\%}&\textcolor[HTML]{0071BB}{-4.8\%}&\textcolor[HTML]{0071BB}{+5\%}&\textcolor[HTML]{0071BB}{-23\%}&\textcolor[HTML]{0071BB}{-3.2\%}&\textcolor[HTML]{0071BB}{+5\%}\\
  
    \bottomrule 
    \label{cmp:arad} 
    \end{tabular}}
\end{table}
\begin{table}[!h]
\centering
    \caption{Performance comparison on the HySpecNet-11k dataset.}
    \renewcommand{\arraystretch}{1.2}
     \resizebox{\textwidth}{!}{
    \begin{tabular}{ccccccccccccccc}
    \toprule 
     \multirow{2}{*}{Models}& \multirow{2}{*}{Params}& \multirow{2}{*}{Times}&\multicolumn{3}{c}{HySpec11k Valid $1.1\times $} & \multicolumn{3}{c}{HySpec11k Valid $0.9\times $}& \multicolumn{3}{c}{HySpec11k Test $1.1\times $} & \multicolumn{3}{c}{HySpec11k Test $0.9\times $} \\ \cline{4-15}
    &&&RMSE & MRAE &PSNR &RMSE &MRAE &PSNR &RMSE &MRAE &PSNR &RMSE &MRAE &PSNR    \\
    \midrule 
    \rowcolor{gray!10}HSCNN+\cite{HSCNN}&7.65M&300.49ms& 0.0335&27.2\% &31.66&0.0376&29.6\%&31.15 &0.0295&21.2\%&32.25&0.0338&24.6\%&31.40\\
    HRNet\cite{hrnet}&31.8M&469.17ms&0.0443&34.2\%&28.74&0.0369&32.3\%&30.19&0.0383&26.3\%&30.20&0.0340&25.0\%&30.76\\

    \rowcolor{gray!10}AGDNet\cite{AGDNet}&6.8M&981.24 ms&0.0326&28.7\%&31.42&0.0361&33.8\%&30.62&0.0282&20.2\%&32.46&0.0339&27.4\%&30.71\\

    GMSR\cite{gmsr}&0.30M&544.42 ms&0.0343&30.7\%&30.95&0.0382&34.8\%&30.23&0.0302&22.0\%&32.04&0.0366&28.5\%&	30.16\\

    \rowcolor{gray!10}MST++\cite{mst++}&1.67M&536.76 ms& 0.0298&27.4\%&32.27&0.0326&30.1\%&31.71&0.0264&201.\%&33.17&0.0312&25.4\%&31.67\\

    R3ST\cite{r3st}&1.79M&592.61 ms&\underline{0.0285}&\underline{26.9\%}&\underline{32.69}&\underline{0.0321}&\underline{30.2\%}&\underline{31.85}&\underline{0.0251}&\underline{19.6\%}&\underline{33.60}&\underline{0.0311}&\underline{25.0\%}&\underline{31.68}\\
    
    \rowcolor{gray!10}MCGA-S2&0.80M&117.41 ms&0.0251&25.4\%&33.34&0.0287&28.7\%&32.42&0.0232&18.5\%&34.08&0.0285&23.1\%&32.43\\

    TTA&0.76M&93.60 ms&\textbf{0.0211}&\textbf{22.1\%}&\textbf{35.06}&\textbf{0.0228}&\textbf{23.7\%}&\textbf{34.37}&\textbf{0.0196}&\textbf{15.1\%}&\textbf{35.78}&\textbf{0.0218}&\textbf{17.2\%}&\textbf{34.87}\\
    
    Improvement&\textcolor[HTML]{0071BB}{-}&\textcolor[HTML]{0071BB}{-}&\textcolor[HTML]{0071BB}{-26\%}&\textcolor[HTML]{0071BB}{-4.8\%}&\textcolor[HTML]{0071BB}{+7\%}&\textcolor[HTML]{0071BB}{-29\%}&\textcolor[HTML]{0071BB}{-6.5\%}&\textcolor[HTML]{0071BB}{+8\%}&
    \textcolor[HTML]{0071BB}{-22\%}
    &\textcolor[HTML]{0071BB}{-4.5\%}&\textcolor[HTML]{0071BB}{+6\%}&\textcolor[HTML]{0071BB}{-30\%}&\textcolor[HTML]{0071BB}{-8\%}&\textcolor[HTML]{0071BB}{+10\%}\\
    \bottomrule 
    \label{cmp:hyspec} 
    \end{tabular}}

\end{table}

\end{document}